\documentclass[letterpaper]{article} 
\usepackage{aaai24}  
\usepackage{times}  
\usepackage{helvet}  
\usepackage{courier}  
\usepackage[hyphens]{url}  
\usepackage{graphicx} 
\usepackage{natbib}  
\usepackage{caption} 
\usepackage{algorithm}
\usepackage{algorithmic}

\usepackage{url}
\usepackage{amsmath}
\usepackage{amsfonts}
\usepackage{multirow}
\usepackage{booktabs}
\usepackage{textcomp}
\usepackage{verbatim}
\usepackage{booktabs}
\usepackage{dsfont}
\usepackage{amssymb}
\usepackage[table,xcdraw]{xcolor}
\usepackage{float}
\usepackage{graphicx}
\usepackage{soul, color, xcolor}

\usepackage{newfloat}
\usepackage{listings}
\DeclareCaptionStyle{ruled}{labelfont=normalfont,labelsep=colon,strut=off} 
\floatstyle{ruled}
\newfloat{listing}{tb}{lst}{}
\floatname{listing}{Listing}
\title{Personalized LoRA for Human-Centered Text Understanding}
\author{
    You Zhang\textsuperscript{\rm 1},
    Jin Wang\textsuperscript{\rm 1}\equalcorreauthor,
    Liang-Chih Yu\textsuperscript{\rm 2}\equalcorreauthor,
    Dan Xu\textsuperscript{\rm 1},
    Xuejie Zhang\textsuperscript{\rm 1}\\
}
\affiliations{
    \textsuperscript{\rm 1}School of Information Science and Engineering, Yunnan University, Yunnan, P.R.China\\
    \textsuperscript{\rm 2}Department of Information Management, Yuan Ze University, Taiwan\\
    \{yzhang0202, wangjin\}@ynu.edu.cn, lcyu@saturn.yzu.edu.tw

}

\usepackage{bibentry}

\begin{document}

\maketitle

\begin{abstract}
Effectively and efficiently adapting a pre-trained language model (PLM) for human-centered text understanding (HCTU) is challenging since user tokens are million-level in most personalized applications and do not have concrete explicit semantics. A standard and parameter-efficient approach (e.g., LoRA) necessitates memorizing numerous suits of adapters for each user. In this work, we introduce a personalized LoRA (PLoRA) with a plug-and-play (PnP) framework for the HCTU task. PLoRA is effective, parameter-efficient, and dynamically deploying in PLMs. Moreover, a personalized dropout and a mutual information maximizing strategies are adopted and hence the proposed PLoRA can be well adapted to few/zero-shot learning scenarios for the cold-start issue. Experiments conducted on four benchmark datasets show that the proposed method outperforms existing methods in full/few/zero-shot learning scenarios for the HCTU task, even though it has fewer trainable parameters. For reproducibility, the code for this paper is available at: 
\url{https://github.com/yoyo-yun/PLoRA}.

\end{abstract}

\section{Introduction} \label{Introduction}
Human-centered text understanding (HCTU) aims to capture potential mental states in texts according to user preferences where user historical written texts are informative sources for user understanding \citep{Crisan2022,Lynn2017,Capel2023}. For users with different preferences and unique needs, similar texts might express various and diverse understandings, e.g., in personalized sentiment analysis \citep{,Zhang2021,Tang2015a,Chen2016}. With the thriving of pre-trained language models (PLMs) in natural language processing (NLP), the importance of personalization has been highlighted in recent works \citep{Wu2023,Min2021,Liu2023}.

Traditional personalized sentiment analysis methods typically use personalized knowledge injection (PKI) techniques \citep{Zhong2021,Houlsby2019,Hu2021} to embed user attributes/preferences into dense representations and then inject them into neural networks, as shown in Fig. 1(a). However, these models require full-model fine-tuning (FFT) and sophisticated structures to couple task-specific transferring and personalized injecting, which is not suitable for adapting large-scale PLMs to personalized sentiment analysis.

To address the problem, parameter-efficient fine-tuning (PEFT) techniques such as the adapter, prompt-tuning, and low-rank adaptation (LoRA) \citep{Wu2018,Zhang2021b} have been proposed. These techniques only require adding and fine-tuning a few parameters in PLMs, and thus can effectively and efficiently fine-tune large-scale PLMs for downstream tasks. To use PEFT for personalized sentiment analysis, as shown in Fig. 1(b), the user attributes/preferences can be considered as an adapter (dashed rectangles) to fine-tune large-scale PLMs through, for example, LoRA, thus avoiding fine-tuning the whole model parameters. However, each user is associated with an adapter and a full model copy is not practicable because user tokens are always million-level in real-world applications (e.g., Amazon). In addition, this may also suffer from the under-fitting problem due to limited training examples for each user.
To leverage the advantage of both PKI and PEFT, this study proposes a personalized low-rank adaptation (PLoRA) mechanism by combining PKI and LoRA, as shown in Fig. 1(c). The PKI is used to embed all user attributes/preferences into a unified embedding, followed by LoRA to adapt large-scale PLMs to the HCTU task without FFT.

\begin{figure*}[t!]
\centering
\includegraphics[width=4.8in]{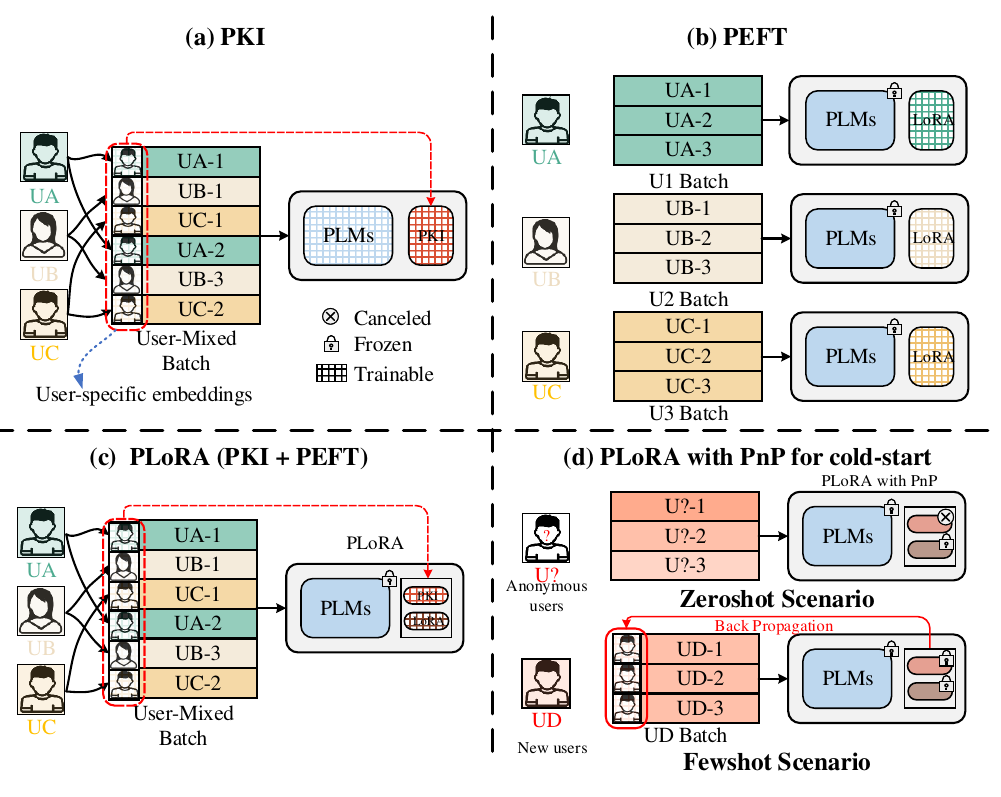}
\centering
\caption{Different methods for human-centered text understanding.}
\label{fig:1}
\end{figure*}

Moreover, we further use the Plug-and-Play (PnP) framework \citep{Sun2022,Zhang2023a} to extend the proposed PLoRA mechanism so that it can be more flexible and easier to be deployed for the cold-start issue including both zero-shot and few-shot learning scenarios \citep{Wu2023,Dathathri2020}, as shown in Fig. 1(d). The zero-shot learning scenario occurs when new and anonymous users due to privacy considerations request a service. In this circumstance, directly applying personalized models to such a zero-shot learning scenario may degrade the prediction performance because the models have no idea about the new users \citep{Zhang2023}.  Conversely, the extended PLoRA can use personalized dropout (PDropout) \citep{10.5555/2627435.2670313} and mutual information maximization (MIM) \citep{NIPS2010_42998cf3} to remove personalized information such that the model’s generalization ability can be increased to handle anonymous new users. The few-shot learning scenario occurs when a new user with a few samples requests a service. Unlike traditional methods that need to retrain the model, both PLoRA and PLMs parameters are frozen. The extended PLoRA is used as a knowledge extractor to produce personalized information for each new user via a backpropagation optimization, and then perform the PnP to complete online training and prediction.

Similar to LoRA, the proposed PLoRA does not increase additional sequence lengths when it handles text input and has no additional inference latency in comparison to other PEFT methods such as adapter \citep{Pfeiffer2020}, pre-fix tuning \citep{Li2021b}, Prompt-tuning \citep{Lester2021}, and P-tuning \citep{Liu2021a}. Moreover, PLoRA is agnostic to neural networks and orthogonal to many prior methods, hence it can be easily deployed in various PLMs and combined with other technologies such as prompt-tuning and prefix-tuning \citep{Houlsby2019,Guo2021}. Experiments conducted on four benchmark datasets show that the proposed method outperforms existing methods in full/few/zero-shot learning scenarios for the HCTU task. The main contributions in this work are as follows.

\begin{itemize}
    \item We proposed the PLoRA mechanism by combining the PKI and LoRA to inject personalized information into PLMs without full-model fine-tuning.
    \item We further extend PLoRA using the PnP framework to increase the model’s generalization ability and enable online training and prediction for the code-start scenario.
    \item Experimental results show that PLoRA is effective, lightweight, easy-to-deploy in PLMs for HCTU tasks.
\end{itemize}

The remainder of this paper is structured as follows. Section 2 describes the details of the proposed method. Extensive experiments are analyzed in Section 3 and conclusions are finally drawn in Section 4. What’s more, the technical appendix provided related work, further proofs, detailed settings and additional experiments.

\section{Methodology} \label{Methodology}
\subsection{Problem Formulation} \label{Methodology-Problem_Formulation}
The HCTU task considers every text data sample belonging to a certain user and captures textual representations according to the user preferences where a collection of written text records is associated with each user. These records facilitate models to understand user preferences and provide personalization services. Consequently, HCTU can be formulated as follows. Given an input text ${\bf{x}} = [{x_1},{x_2},...,{x_N}]$, the goal of human-centered models is to generate an output $y$ for the user $u$, where $N$ represents the input length and $y$ associates task targets such as sentiment scores. These tasks can be modeled as $\arg \max q(y|{\bf{x}},u)$. For each user $u$, a collection of data is denoted as $ {{{\mathcal D}}_u} = \sum\nolimits_i {({{\bf{x}}_i},{y_i},u)} $ on which user preference can be generated.

To simulate cold-start issues including zero-shot and few-shot learning scenarios, a couple of data (${{{\mathcal D}}^{\rm{A}}}$ and ${{{\mathcal D}}^{\rm{B}}}$) are provided for each dataset ${{\mathcal D}} = \sum\nolimits_u {\sum\nolimits_i {({{\bf{x}}_i},{y_i},u)} }  = {{{\mathcal D}}^{\rm{A}}} + {{{\mathcal D}}^{\rm{B}}}$ where $(\forall {u^{\rm{A}}} \in {{{\mathcal D}}^{\rm{A}}}) \cap (\forall {u^{\rm{B}}} \in {{{\mathcal D}}^{\rm{B}}}) = \emptyset $. ${{{\mathcal D}}^{\rm{A}}}$ is used for full-shot learning and ${{{\mathcal D}}^{\rm{B}}}$ is provided for cold-start evaluation. For zero-shot learning, it requires models $q(y|{\bf{x}},u)$ learned from ${{{\mathcal D}}^{\rm{A}}}$ to show superior generalization performance $q(y|{\bf{x}})$ on $({\bf{x}},y) \in {{{\mathcal D}}^{\rm{B}}}$ without user $u^{\rm{B}}$ as input. For few-shot learning, the user preference $p = f(u)$ is learned from a few data ${{{\mathcal D}}_u}$ and then used for personalization services $q(y|{\bf{x}},u)$ where $({\bf{x}},y,u) \in {{{\mathcal D}}_u}$ and ${{{\mathcal D}}_u} \in {{{\mathcal D}}^{\rm{B}}}$.

\subsection{Personalized LoRA} \label{Methodology-Personalized_LoRA}
In this section, we elaborate on PLoRA for personalization in PLMs, as illustrated in Figure 2, which combines task-specific LoRA and user-specific PKI.

\begin{figure}[t!]
\centering
\includegraphics[width=3.0in]{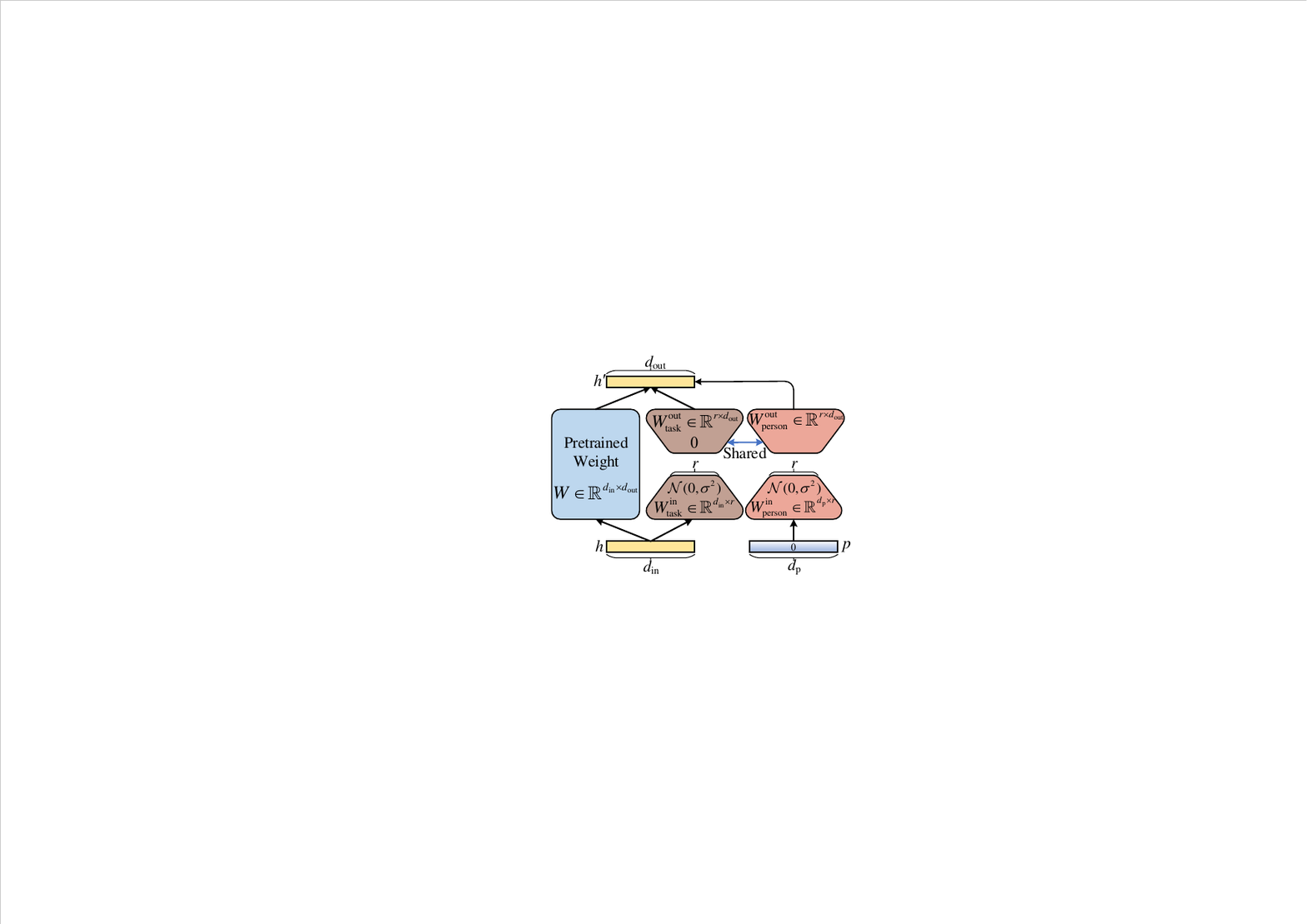}
\centering
\caption{An illustration of PLoRA.}
\label{fig:2}
\end{figure}

\subsubsection{Task-specific LoRA} \label{LoRA}
A neural network such as PLMs contains many matrix multiplications. LoRA provides a PEFT that facilitates weight matrices to capture intrinsic rank for downstream tasks adaptation. For a weight matrix $W \in {\mathbb{R}^{{d_{{\rm{in}}}} \times {d_{{\rm{out}}}}}}$ in PLM, lightweight matrices of $W_{{\rm{task}}}^{{\rm{in}}} \in {\mathbb{R}^{{d_{{\rm{in}}}} \times r}}$ and $W_{{\rm{task}}}^{{\rm{out}}} \in {\mathbb{R}^{r \times {d_{{\rm{out}}}}}}$ with a low rank of $r \ll \min ({d_{{\rm{in}}}},{d_{{\rm{out}}}})$ are additionally applied to update $W$ as $W + W_{{\rm{task}}}^{{\rm{in}}}W_{{\rm{task}}}^{{\rm{out}}}$. Given $h \in {\mathbb{R}^{{d_{{\rm{in}}}}}}$ as input textual representation, the output $h' \in {\mathbb{R}^{{d_{{\rm{out}}}}}}$ is calculated with task-specific LoRA via:
\begin{equation} \label{eq:1}
h' = hW + hW_{{\rm{task}}}^{{\rm{in}}}W_{{\rm{task}}}^{{\rm{out}}} \,.
\end{equation}
In the training phase, $W$ is fixed and only $W_{{\rm{task}}}^{{\rm{in}}}$ and $W_{{\rm{task}}}^{{\rm{out}}}$ are learned.

\subsubsection{Personalization-specific PKI} \label{PKI}
To adapt a generic model to personalization, most of the existing works are dedicated to incorporating personalized knowledge $p = f(u) \in {\mathbb{R}^{{d_p}}}$ with textual representation via:
\begin{equation} \label{eq:2}
h' = hW + p{W_{{\rm{person}}}} \,,
\end{equation}
where $f(u)$ means user preferences or embeddings of $u$. Unfortunately, such methods require an optimization of both $W$ and $W_{\rm {person}}$ in an FFT method so as to coordinate with personalization and downstream tasks, making their deployment cumbersome in practice.

\subsubsection{PLoRA} \label{PLoRA}
 To compose a parameter-efficient and personalized adapter in PLMs, PLoRA takes $h$ and $p$ as inputs and outputs $h'$ via:
\begin{equation} \label{eq:3}
\begin{aligned}
  h' &= hW + hW_{{\rm{task}}}^{{\rm{in}}}W_{{\rm{task}}}^{{\rm{out}}} + p{W_{{\rm{person}}}} \cr
     &= hW + hW_{{\rm{task}}}^{{\rm{in}}}W_{{\rm{task}}}^{{\rm{out}}} + pW_{{\rm{person}}}^{{\rm{in}}}W_{{\rm{person}}}^{{\rm{out}}} \cr
     &= hW + (hW_{{\rm{task}}}^{{\rm{in}}} + pW_{{\rm{person}}}^{{\rm{in}}})W_{{\rm{task}}}^{{\rm{out}}} \cr
\end{aligned} \,.
\end{equation}
Initially, a low-rank hypothesis is applied to PKI with ${W_{{\rm{person}}}} = W_{{\rm{person}}}^{{\rm{in}}}W_{{\rm{person}}}^{{\rm{out}}}$ for capture intrinsic rank features of injection weights. To further facilitate the couples with task-specific adaptation and personalization, we share $W_{{\rm{task}}}^{{\rm{out}}}$ with $W_{{\rm{person}}}^{{\rm{out}}}$. Similar to LoRA, only $W_{{\rm{task}}}^{{\rm{in}}}$, $W_{{\rm{task}}}^{{\rm{out}}}$ (or $W_{{\rm{person}}}^{{\rm{out}}}$), and $W_{{\rm{person}}}^{{\rm{in}}}$ are trainable parameters that receive gradient updates.

\begin{figure}[t!]
\centering
\includegraphics[width=3.0in]{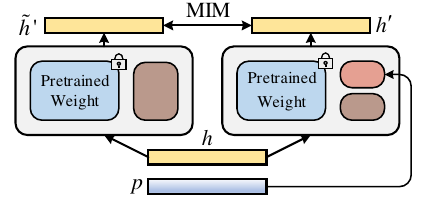}
\centering
\caption{A diagram of MIM on PLoRA for zero-shot learn-ing.}
\label{fig:3}
\end{figure}

\subsection{Plug-and-Play for Cold Start} \label{Methodology-Plug-and-Play_for_Cold_Start}
Large-scale PLMs trained on a large number of corpora show excellent text-understanding capabilities. Instead of FFT or modifying model architecture for task-specific domains, the LoRA module can be regarded as a PnP module that guides PLMs to be adapted to target domains such as sentiment analysis. Similarly, PKI can also be considered as a PnP module that guides PLMs to serve different users. Note that, when PnP modules are applied, PLMs will not be trained and thus their architecture will not be modified and their pre-training power will not be deteriorated.

To build a PnP framework and apply PLoRA for complex cold-start issues, a reparameterization strategy is proposed. Especially, $W_{{\rm{task}}}^{{\rm{in}}}$ and $W_{{\rm{person}}}^{{\rm{in}}}$ are randomly in Gaussian distributions; $W_{{\rm{task}}}^{{\rm{out}}}$ and $p$ are zeros. Hence, $W_{{\rm{task}}}^{{\rm{in}}}W_{{\rm{task}}}^{{\rm{out}}}$ is zero at the beginning of the training phase and $pW_{{\rm{person}}}^{{\rm{in}}}W_{{\rm{person}}}^{{\rm{out}}}$ is zero when PLMs first meet a user $u$. As a result, PLoRA is a PnP module that is orthogonal to PLMs and gradually grows with capabilities of guiding PLMs to be adapted to task-specific and human-centered domains.

Although high-performance PLoRA gains, it is still difficult for zero-shot learning and few-shot learning scenarios (See Section Experiments). 

\subsubsection{Zero-shot learning scenarios} \label{zeroshot}
It is challenging in zero-shot learning since poor decomposition of generic and human-centered features makes PLoRA degrading performance for unseen or anonymous users. To address this problem, we propose PDropout and MIM methods to help PLoRA to remain its generalization performance for zero-shot learning scenarios.

\begin{figure}[t!]
\centering
\includegraphics[width=3.0in]{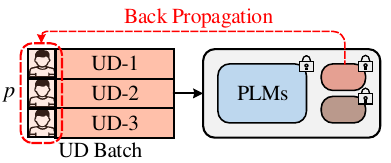}
\centering
\caption{A diagram of PnP framework on few-shot learning scenarios for user D (UD).}
\label{fig:4}
\end{figure}

\begin{itemize}
    \item {\bf PDropout}. During the training, we randomly mask out users in a batch of samples with a dropout ratio of $\omega  \in [0,1]$. Consequently, with the capability of adapting to personalization, the PLoRA framework remains accessible to generic task performance.
    \item {\bf Mutual Information Maximization}. Inspired by \citep{Zhang2023}, we introduce an MIM method to align the distance between task-specific and human-centered representations ($\tilde h'$ and $h'$), as shown in Figure 3. During the training, human-centered representation regarded as a teacher states guides better generic task-specific performance where MIM can be considered as a knowledge distillation mechanism. By contrast, MIM is also a regularization term to leverage generalization performance on human-centered representation. In practice, both mean square error (MSE) and Kullback-Leibler (KL) divergence~\citep{kullback1951information} are prevalent for instantiating MIM.
\end{itemize}

\subsubsection{Few-shot learning scenarios} \label{fewshot}
For few-shot learning as shown in Figure 4, we fix all model parameters and inject a zeroed embedding $p$ referring to a new user $u$. With a few samples associated with the user, HCTU models update the user preference $p$ through a backpropagation mechanism so that the user could participate in personalized service with updated $p$ as an access token. 

\subsection{Train and Inference} \label{Methodology-Train_and_Inference}
In principle, the proposed PLoRA can be applied to any subset of linear projectors in neural networks to provide PEFT and personalization. In this work, we limit our method to multi-head attention mechanisms (i.e., query and value projectors) in the Transformer structure for downstream tasks, following LoRA study \citep{Hu2021}.

For optimization in the training phase, the loss function of full-shot learning is defined as follows:
\begin{equation} \label{eq:4}
\begin{aligned}
  {{{\cal L}}_{{\rm{Fullshot}}}} &= {{{\cal L}}_{{\rm{CE}}}} + \alpha {{{\cal L}}_{{\rm{MIM}}}} \cr 
  {{{\cal L}}_{{\rm{CE}}}} &= {\rm{CE}}(q(\tilde y|{\bf{x}},{\rm{PDropout}}(u,\omega );\theta ),y) \cr 
  {{{\cal L}}_{{\rm{MIM}}}} &= \sum\limits_{(\tilde h',h') \in q(\tilde y|{\bf{x}},u)} {{\rm{MIM}}(\tilde h',h')} \cr
\end{aligned}\,,
\end{equation}
where $({\bf{x}},y,u) \in {{{\cal D}}^{\rm{A}}}$ refers to each sample; ${\rm{PDropout}}( \cdot )$ and ${\rm{MIM}}( \cdot )$ are the proposed PDropout and MIM methods; $\theta $ presents lightweight trainable parameters; $\alpha $ is balance ratio for MIM term.

For few-shot learning, the training objective is formulated as:
\begin{equation} \label{eq:5}
\begin{aligned}
f(u) & = \mathop {\arg \min }\limits_{p = f(u)} {{{\cal L}}_{{\rm{Fewshot}}}} \cr 
  {{{\cal L}}_{{\rm{Fewshot}}}}  &= \sum\limits_{({\bf{x}},y,u) \in {{{\cal D}}^{\rm{B}}}} {{\rm{CE}}(q(\tilde y|{\bf{x}},u;\hat \theta ),y)}  \cr
\end{aligned}\,,
\end{equation}
where $\hat \theta $ presents well-trained parameters in full-shot learning scenarios and updated $f(u)$ stand for the semantics of user $u$.

Similar to LoRA, the proposed PLoRA still has no additional inference latency in deployment. We can explicitly compute $W = W + W_{{\rm{task}}}^{{\rm{in}}}W_{{\rm{task}}}^{{\rm{out}}}$ and $b = b + f(u) \cdot W_{{\rm{person}}}^{{\rm{in}}}W_{{\rm{person}}}^{{\rm{out}}}$ to update a linear projector for personalized services for the user $u$ where the linear projector contains weight matrix $W$ and bias vector $b$. When we need to switch to pure (or generic) task-specific tasks or anonymous users, we can recover $b$ by subtracting $f(u) \cdot W_{{\rm{person}}}^{{\rm{in}}}W_{{\rm{person}}}^{{\rm{out}}}$. Furthermore, we can also add different $f(u') \cdot W_{{\rm{person}}}^{{\rm{in}}}W_{{\rm{person}}}^{{\rm{out}}}$ terms to switch services for the user $u'$. As a consequence, PLoRA can also switch to other tasks with very little memory overhead.

\begin{table*}[t]
\centering
\small
\begin{tabular}{|c|c|c|c|c|c|c|c|c|c|c|c|c|c|} \hline
\multirow{2}{*}{\textbf{Datasets}} & \multicolumn{3}{c|}{IMDB-A}                     & \multicolumn{3}{c|}{YELP-A}                     & \multicolumn{3}{c|}{GDRD-A}                     & \multicolumn{3}{c|}{PPR-A}                      & \multirow{2}{*}{TP(\%)} \\\cline{2-13}
                          & $Acc$           & $MSE$            & $F_1$            & Acc           & $MSE$            & $F_1$            & $Acc$           & $MSE$            & $F_1$            & $Acc$           & $MSE$            & $F_1$            &                         \\\hline
\multicolumn{14}{|c|}{Generic sentiment analysis}                                                                                                                                                                                                         \\\hline
NSC                       & 47.7          & 1.729          & 44.7          & 65.2          & 0.452          & 62.6          & -             & -              & -             & -             & -              & -             & 100x                    \\
BERT                      & \underline{51.7}    & 1.398          & \underline{50.3}    & 67.6          & \underline{0.390}    & 65.1          & \underline{57.0}    & \underline{0.582}    & \underline{53.7}    & \underline{83.1}    & \underline{0.192}    & 45.7          & 100x                    \\
$\cal B$-LoRA                    & 51.0          & \underline{1.349}    & 48.9          & \underline{67.8}    & 0.400          & \underline{66.2}    & 56.8          & 0.590          & 53.2          & 82.6          & 0.198          & \underline{47.8}    & 2.1x                    \\ \hline
$\cal R$-LoRA                    & \textbf{52.5} & 1.252          & 50.5          & 69.6          & 0.354          & 68.9          & 58.4          & 0.566          & 56.0          & 83.6          & 0.188          & \textbf{55.5} & 2.3x                    \\
$\cal FT$5-LoRA                  & \textbf{52.5} & \textbf{1.218} & \textbf{50.8} & \textbf{70.0} & \textbf{0.352} & \textbf{69.3} & \textbf{58.1} & \textbf{0.553} & \textbf{55.9} & \textbf{84.5} & \textbf{0.168} & 53.8          & 2.7x                    \\\hline
\multicolumn{14}{|c|}{Human-centered sentiment analysis}                                                                                                                                                                                                  \\\hline
NSC+U                     & 51.1          & 1.460          & 47.1          & 67.3          & 0.444          & 64.9         & -             & -              & -             & -             & -              & -             & 100x                    \\
$\cal B$-MAA                     & 55.4          & \underline{1.129}    & 53.4          & 71.6          & 0.352          & 69.8          & 63.6          & 0.427          & 58.3          & 84.5          & 0.173          & 41.5          & 150x                    \\
$\cal B$-PKI                     & 55.6          & 1.200          & 53.2          & 70.1          & 0.377          & 68.5          & 64.8          & 0.470          & 60.8          & 84.7          & 0.173          & 48.3          & 2.1x                    \\
$\cal B$-UserAdapter             & 55.7          & 1.131          & 53.5          & 70.8          & 0.357          & 68.5          & 64.7          & 0.465          & 60.0          & 84.4          & 0.175          & 48.0          & 100x                    \\
$\cal B$-PLoRA$_{2S}$                 & 54.5          & 1.246          & 52.0          & 70.7          & 0.362          & 68.9          & 63.9          & 0.493          & 59.9          & 84.5          & 0.174          & 47.8          & 3.1x                    \\
$\cal B$-PLoRA                   & \underline{57.0}    & 1.151          & \underline{54.9}    & \underline{72.1}    & \underline{0.338}    & \underline{70.3}    & \underline{65.0}    & \underline{0.463}    & \underline{61.4}    & \underline{85.1}    & \underline{0.168}    & \underline{49.8}    & 3.1x                    \\ \hline
$\cal R$-PLoRA                   & 58.0          & 1.008          & 55.8          & 72.9          & 0.318          & 71.3          & 65.8          & 0.457          & 61.8          & 85.5          & 0.158          & \textbf{64.1} & 3.2x                    \\
$\cal FT$5-PLoRA                 & \textbf{58.7} & \textbf{0.980} & \textbf{56.1} & \textbf{73.3} & \textbf{0.314} & \textbf{71.5} & \textbf{66.7} & \textbf{0.447} & \textbf{63.3} & \textbf{86.7} & \textbf{0.144} & 60.4          & 3.8x     \\\hline              
\end{tabular}
\caption{Comparative test results on full-shot learning scenarios. TP presents a trainable parameter (including user embeddings) ratio during the optimization. All figures are averaged over five runs. The underscored and black-face figures mean the best scores for only $\cal B$ and all experiments in each group, respectively.} \label{tab:1}
\end{table*}

\begin{table*}[t]
\centering
\small
\begin{tabular}{|c|c|c|c|c|c|c|c|c|c|c|c|c|c|} \hline
\multirow{2}{*}{\textbf{Datasets}} & \multicolumn{3}{c|}{IMDB-B}                     & \multicolumn{3}{c|}{YELP-B}                     & \multicolumn{3}{c|}{GDRD-B}                                & \multicolumn{3}{c|}{PPR-B}                      & \multirow{2}{*}{TP(\%)} \\ \cline{2-13}
                          & $Acc$           & $MSE$            & $F_1$            & $Acc$           & $MSE$            & $F_1$            & $Acc$                 & $MSE$                  & $F_1$            & $Acc$           & $MSE$            & $F_1$            &                         \\\hline
\multicolumn{14}{|c|}{Few-shot learning}                                                                                                                                                                                                                              \\\hline
$\cal B$-PLoRA*                  & 50.2          & 1.722          & 47.3          & 67.1          & 0.479          & 64.8          & 58.5                & 0.632                & 54.8          & 83.7          & 0.188          & 44.3          & 2.9x                    \\
$\cal B$-MAA (FS)                & 54.0          & 1.652          & 51.8          & 71.0          & 0.379          & 68.9          & 63.8                & 0.535                & 59.4          & 84.0          & 0.188          & 40.2          & 0.2x                    \\
$\cal B$-UserAdapter (FS)        & 52.9          & \underline{1.471}    & 50.9          & 70.9          & 0.383          & 69.2          & \underline{\textbf{65.2}} & \underline{\textbf{0.502}} & 60.0          & 83.8          & 0.168          & \underline{44.8}    & 0.2x                    \\
$\cal B$-PLoRA$_{2S}$ (FS)            & 54.2          & 1.641          & 52.1          & 70.1          & 0.410          & 68.1          & 61.6                & 0.611                & 57.1          & 84.1          & 0.168          & 42.6          & 0.1x                    \\
$\cal B$-PLoRA (FS)              & \underline{55.1}    & 1.559          & \underline{53.5}    & \underline{71.2}    & \underline{0.362}    & \underline{69.3}    & 64.0                & 0.536                & \underline{\textbf{60.5}} & \underline{84.5}    & \underline{0.164}    & 43.7          & 0.1x                    \\ \hline
$\cal R$-PLoRA (FS)              & 56.5          & 1.541          & 54.7          & 72.6          & 0.355          & 71.3          & 63.9                & 0.528                & 59.8          & 84.7          & 0.171          & 57.8          & 0.1x                    \\
$\cal FT$5-PLoRA (FS)            & \textbf{58.0} & \textbf{1.396} & \textbf{55.8} & \textbf{73.2} & \textbf{0.338} & \textbf{72.3} & 63.1                & 0.514                & 59.6          & \textbf{86.1} & \textbf{0.148} & \textbf{59.4} & 0.1x                    \\\hline
\multicolumn{14}{|c|}{Zero-shot learning}                                                                                                                                                                                                                             \\\hline
$\cal B$-LoRA*                   & \underline{47.2}    & 2.232          & 44.4          & 65.0          & 0.497          & 62.7          & 52.3                & 0.687                & 48.3          & 81.4          & 0.218          & 34.5          & 2.1x                    \\
$\cal B$-UserAdapter             & 44.3          & 2.123          & 41.9          & 65.6          & 0.418          & 65.1          & 53.2                & 0.681                & 51.5          & 82.1          & 0.185          & \underline{43.6}          & -                       \\
$\cal B$-PLoRA$_{2S}$                 & 44.1          & 1.889          & 41.6          & \underline{67.3}    & 0.416          & 65.4          & 53.5                & 0.676                & 51.5          & 82.3          & 0.183    & 41.5          & -                       \\
$\cal B$-PLoRA (ZS)              & 46.9          & \underline{1.632}    & \underline{45.1}    & \underline{67.3}    & \underline{0.412}    & \underline{65.8}    & \underline{56.2}          & \underline{0.640}          & \underline{53.7}    & \underline{83.3}    & \underline{0.180}          & 42.8          & -                       \\\hline
$\cal R$-PLoRA (ZS)              & 47.6          & 1.553          & 46.0          & 68.6          & 0.383          & 67.5          & \textbf{56.6}       & 0.630                & \textbf{55.4} & 83.1          & 0.187          & 53.7          & -                       \\
$\cal FT$5-PLoRA (ZS)            & \textbf{48.1} & \textbf{1.521} & \textbf{46.5} & \textbf{70.7} & \textbf{0.352} & \textbf{68.8} & 54.3                & \textbf{0.618}       & 52.8          & \textbf{84.7} & \textbf{0.159} & \textbf{57.4} & -   \\\hline                   
\end{tabular}
\caption{Comparative test results on few/zero-shot learning scenarios. * presents corresponding models optimized only from ${{{\cal D}}^{\rm{B}}}$. FS means corresponding models are first learned in ${{{\cal D}}^{\rm{A}}}$ and are then adapted to ${{{\cal D}}^{\rm{B}}}$ in a few-shot learning strategy, i.e., Eq (5). ZS denotes corresponding models optimized with PDropout or MIM in ${{{\cal D}}^{\rm{A}}}$ are directly applied for ${{{\cal D}}^{\rm{B}}}$ without user inputs.}\label{tab:2}
\end{table*}

\section{Experiments} \label{Experiments}
To investigate the effectiveness and efficiency of the proposed methods for HCTU, extensive experiments were conducted and analyzed on personalized sentiment analysis.

\subsection{Datasets and Evaluation} \label{Experiments-Datasets_and_Evaluation}
\subsubsection{Datasets}
We took personalized sentiment analysis as a test bed since sentiment is a critical and sensitive evaluation for human subjective expression. The used datasets include IMDB, YELP, GDRD, and PPR, where a collection of data is associated with different users. To simulate complex and practical situations in real-world applications, all datasets are individually divided into two parts ${{{\cal D}}^{\rm{A}}}$ and ${{{\cal D}}^{\rm{B}}}$ where ${{{\cal D}}^{\rm{A}}}$ contains much larger samples than ${{{\cal D}}^{\rm{B}}}$ and ${{{\cal D}}^{\rm{B}}}$ aims to simulate cold-start scenarios, as formulated in Section Methodology. Either ${{{\cal D}}^{\rm{A}}}$ or ${{{\cal D}}^{\rm{B}}}$, it splits into train, dev, and test data for experiments. More detailed statistics of datasets were listed in Appendices.

\subsubsection{Evaluation}
To measure the effectiveness, Accuracy ($Acc$), mean squared error ($MSE$), and macro $F_1$ ($F_1$) were adopted as evaluation metrics. Note that, higher $F_1$ and $Acc$ (\%) and lower $MSE$ mean better results.

\subsection{Experimental Setup} \label{Experiments-Experimental_Setup}
To perform personalized sentiment analysis using PLMs, we applied PLoRA to multi-head attention including query and value projectors, following \citep{Hu2021}, where introduced PLMs involved BERT ($\cal B$) \citep{Devlin2019}, RoBERT ($\cal R$) \citep{Liu2019}, and Flan-T5 (${\cal FT}5$) \citep{Chung2022}. For the reproduction of experiments, more implementation details of hyperparameters were reported in Appendices.

We compare our methods with the previous high-performance models, including neural sentiment classification (NSC) \citep{Chen2016}, multi-attribute attention (MAA) from MA-BERT \citep{Zhang2021}, and UserAdatper \citep{Zhong2021}. Moreover, several comparative methods derived from our motivations were adopted, including PKI (only PKI tuned in optimization), LoRA, and two-stage (2S). Note that 2S means human-centered models were learned from generic data in advance and then adapted to various users via the few-shot learning strategy in Eq. (5) so that updated models were well capable of providing generic and personalized services.

\begin{table*}[t]
\centering
\begin{tabular}{|l|c|c|c|c|c|c|} \hline
\multicolumn{1}{|c|}{\multirow{2}{*}{\textbf{Models}}} & IMDB-A    & \multicolumn{2}{c|}{IMDB-B} & YELP-A    & \multicolumn{2}{c|}{YELP-B} \\ \cline{2-7}
\multicolumn{1}{|c|}{}                        & Full-shot & Few-shot    & Zero-shot    & Full-shot & Few-shot    & Zero-shot    \\ \hline
$\cal B$-PLoRA                                     & 54.9      & 53.5        & 45.1         & 70.3      & 69.3        & 66.1         \\ \hline
\quad w/o PKI                                     & 48.9      & -           & 41.6         & 66.2      & -           & 65.4         \\ 
\quad w/o LoRA                                    & 53.2      & 52.0        & -            & 68.5      & 67.7        & -            \\ \hline
\quad w/o PDropout                                & 54.2      & 51.4        & 45.0         & 69.8      & 68.9        & 65.7         \\
\quad w/o MIM                                     & 53.9      & 51.7        & 44.0         & 69.4      & 68.6        & 65.8         \\
\quad w/o PDropout \& MIM                         & 54.2      & 51.7        & 41.9         & 69.2      & 68.1        & 59.9        \\ \hline
\end{tabular}
\caption{Ablation study of test $F_1$ scores on IMDB and YELP with respect to full/few/zero-shot learning scenarios.} \label{tab:3}
\end{table*}

\subsection{Comparative Results and Analysis} \label{Experiments-Comparative_Results_and_Analysis}
\subsubsection{Full-shot learning}
Table 1 reported comparative results for both generic (the first group) and human-centered (the second group) scenarios in ${{{\cal D}}^{\rm{A}}}$. From the first group, applying the LoRA method to PLMs facilitated general language knowledge learned from large corpora to be adapted to sentiment analysis tasks with a few trainable parameters. Accordingly, LoRA-based PLMs achieved comparable performance in comparison to FFT models.

Against the first group, models in the second group gained better results on three metrics. This is because the introduction of personalized knowledge helps these models to accurately locate users’ implicate sentiments where sentiment preferences from different users might differ. Based on the same PLMs, i.e., BERT, the proposed method of $\cal B$-PLoRA was on par with previous best-performance models, i.e., $\cal B$-MAA and $\cal B$-UserAdapter concerning their effectiveness. However, MAA and UserAdapter necessitate FFT optimization for task adaptations. Moreover, $\cal B$-PKI showed relatively lower scores, indicating FFT is essential for PLM adapted to downstream tasks. By contrast, the proposed model is more efficient due to the dynamic combination between LoRA and PKI. Not only encoder-based PLMs, but decoder-based T5 can load PLoRA for downstream tasks, revealing an easy-to-deploy capability of PLoRA on wide applications.

In real-world applications, cold-start issues are serious in personalized services due to user-specific domain adaptation and data sparsity. We simulated cold-start issues as few-shot learning and zero-shot learning scenarios and conducted corresponding experiments, as reported in Table 2.

\subsubsection{Few-shot Learning}
To make human-centered models capable of serving on unseen users where the users were from ${{{\cal D}}^{\rm{B}}}$ while out of ${{{\cal D}}^{\rm{A}}}$, we tested the proposed PLoRA and previous works via the introduced few-shot learning strategy from Eq. (5). From the table, it can be found that these models outperformed $\cal B$-LoRA* that was directly optimized on datasets ${{{\cal D}}^{\rm{B}}}$ in a full-shot way since these models updated from ${{{\cal D}}^{\rm{A}}}$ in advance could store robust performance and then be easily deployed to unseen users with a few samples.

\subsubsection{Zero-shot Learning}
Theoretically, PKI-based methods with sophisticated structures, i.e., NSC+U and $\cal B$-MAA, were hard to directly handled pure textual data as they were difficult to distill pure textual representation from human-centered features \citep{Zhang2023}. Fortunately, UserAdatper could directly discard user tokens at input layer and performed well when they met unseen users in zero-shot learning scenarios. $\cal B$-PLoRA$_{2S}$ separately and step-by-step optimized task-specific and use-specific plugins so that it competitively performed in both zero-shot learning and few-shot learning scenarios. However, lacking mutual learning between LoRA (for task adaptation) and PKI (for personalization) made $\cal B$-PLoRA$_{2S}$ method underperform our proposed PLoRA (ZS) that applies PDropout and MIM strategies to decouple task-specific and user-specific knowledge in the full-shot learning scenarios.

\begin{figure}[t!]
\centering
\includegraphics[width=3.3in]{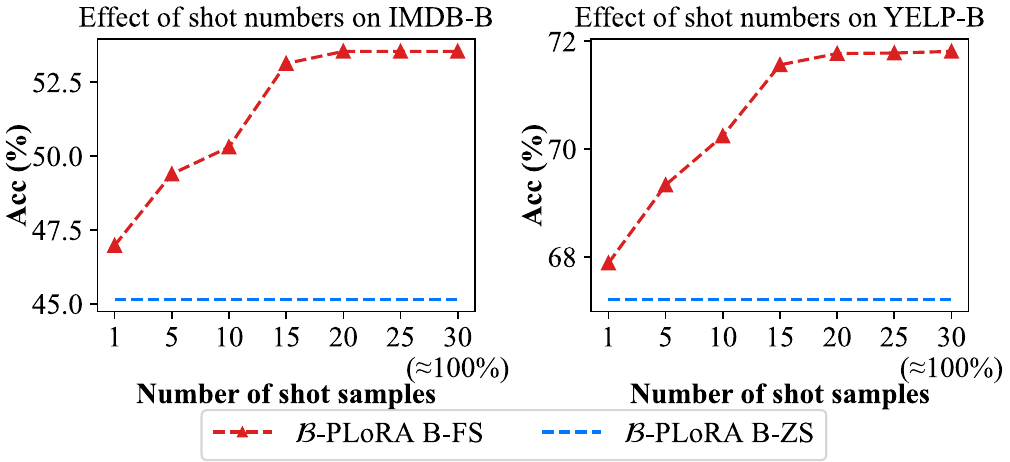}
\centering
\caption{Effect of PLoRA with different data sparsity. FuS, FS, and ZS in figures (including the following figures) means using full/few/zero-shot learning methods, respectively. \{$\cal M$\}-PLoRA \{A/B\}-\{FuS/ZS/FS\} corresponds dev figures of PLM $\cal M$-based PLoRA applied for datasets (A or B) with FuS/ZS/FS methods. $ \approx 100\% $ means almost the full training dataset in ${{{\cal D}}^{\rm{B}}}$ is used and also presents no data sparsity for every user in ${{{\cal D}}^{\rm{B}}}$.}
\label{fig:5}
\end{figure}

\subsection{Model Analysis} \label{Experiments-Model_Analysis}
\subsubsection{Ablation study} \label{Ablation study}
Results of an ablation study on IMDB and YELP were reported in Table 3, where the ablation target included: 1) PKI and LoRA to investigate the effectiveness of PLoRA in task-specific and user-specific knowledge fusion; 2) PDropout and MIM to validate the PnP performance of knowledge decoupling.
From the first group in the table, the performance of PLoRA degraded with the elimination of PKI or LoRA, demonstrating that both task-specific and user-specific adaptations were crucial for the deployment of PLMs in HCTU applications.

From the last group, it can be found that, without PDropout or MIM, PLoRA slightly dropped $F_1$ score in all three scenarios; while it performed catastrophic descents in zero-shot learning scenarios when neither PDropout nor MIM was adopted. This is because PLoRA have coupled task-specific and user-specific knowledge in the full-shot learning procedure, leading to inferior performance of uncoupling task-specific knowledge for generic sentiment analysis if without zero-shot learning strategies. This phenomenon also indicated that either PDropout or MIM could facilitate the proposed PLoRA effectively adapted to generic scenarios. Note that, a dynamic combination of PDropout and MIM is beneficial for better adaptation.

\subsubsection{Sparsity}  \label{Sparsity}
To reveal the robustness of PLoRA, we explored its dataset B-dev $Acc$ in few-shot learning scenarios with different degrees of data sparsity or the number of few-shot samples in B-training sets, as shown in Figure 5. It can be found that the improvements of PLoRA increased with the number of shot samples increasing or the data sparsity degrees decreasing. For both our used IMDB and YELP datasets, 15-shot learning relatively gained competitive dev performance. It indicated that, with sufficient user-oriented data, the proposed method could extract user-specific knowledge which then helps sentiment models to be adapted to such users in turn. Even in a few samples used (such as 1, 5, and 10), PLoRAs with the few-shot learning strategy (i.e., $\cal B$-PLoRA B-FS) still outperformed those (i.e., $\cal B$-PLoRA B-ZS) that were directly applied to zero-shot learning scenarios, indicating the effect of the few-shot learning strategy.

\subsubsection{The sensitivity of hyperparameters.} \label{sensitivity}
To further investigate how each component of PLoRA affects the final performance, we conducted several experiments on the sensitivity of hyperparameters, as shown in Figure 6-7.

Among the four subfigures in Figure 6, lower rank dimensionalities of applied weights and lower user embeddings for user semantics would not support complete performance in all scenarios. By contrast, larger figures of PLoRA configuration might saturate model performance and desire exponential-enlarged trainable parameters. It indicated that, in practice, empirical explorations of an appropriate PLoRA configuration and considerations of limited compute budget were critical for satisfactory services.

To further investigate how PDropout and MIM affect the effectiveness in zero-shot learning scenarios, we conducted analytical experiments in Figure 7. From the figure, it can be first found that, without PDropout nor MIM ($\omega  = \alpha  = 0$), $\cal B$-PLoRA B-ZS did not perform well in zero-shot scenarios, even worse than $\cal B$-LoRA B-ZS (green dashed lines or $\omega  = 1$) that was learned from dataset ${{{\cal D}}^{\rm{A}}}$ without PKI and was then directed applied for dataset ${{{\cal D}}^{\rm{B}}}$ in zero-shot scenarios. Secondly, with appropriate configurations of $\omega $ and $\alpha $, $\cal B$-PLoRA could effectively mitigate the above degradation, in consistent with the analysis in Table 3.

\begin{figure}[t!]
\centering
\includegraphics[width=3.3in]{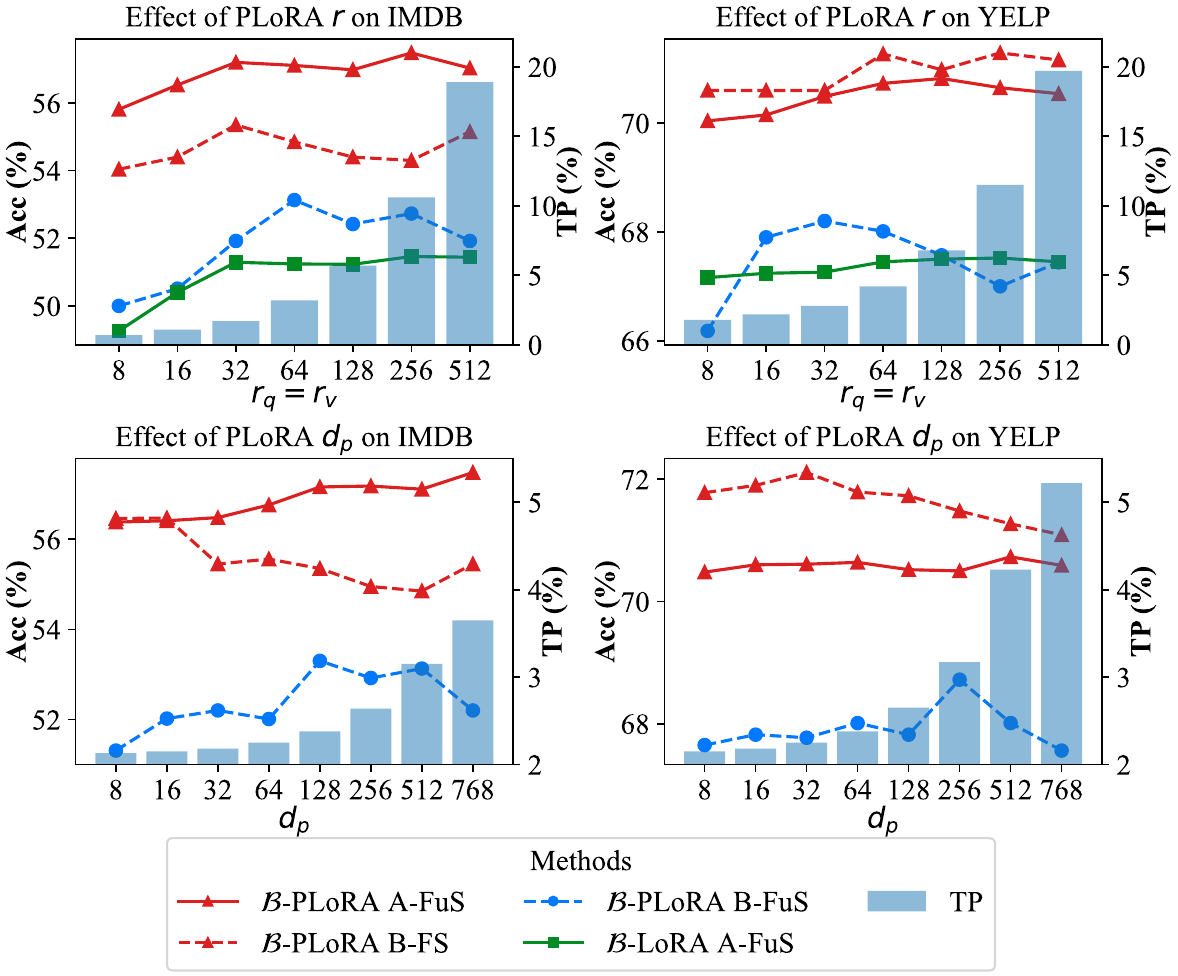}
\centering
\caption{Performance on dev datasets with diverse parameters of PLoRA.}
\label{fig:6}
\end{figure}

\begin{figure}[t!]
\centering
\includegraphics[width=3.3in]{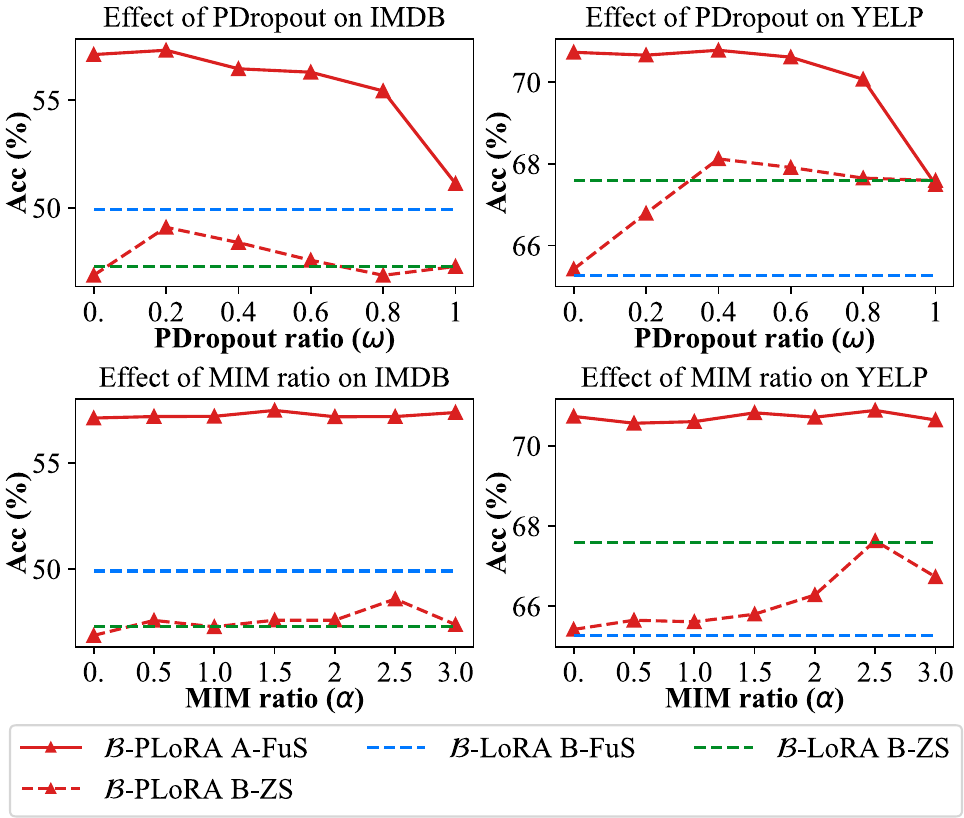}
\centering
\caption{Performance on dev datasets with diverse parameters of PDropout and MIM. Note that, one of them is investi-gated with the other being not applied.}
\label{fig:7}
\end{figure}

\subsection{Discussions} \label{Experiments-Discussions}
In summary, the priority of the proposed PLoRA was located at 1) a dynamic combination of PEFT-based LoRA and knowledge-injected PKI for task-specific and human-centered adaptation; 2) an introduction of the few-shot learning strategy to adapt well-trained PLMs to unseen users in the training phase; 3) a PnP framework that couples and decouples task-specific and human-centered knowledge for cold-start issues.

We conducted experiments with PLMs including BERT, RoBERTa, and Flan-T5, which does not rashly verify that PLoRA could be only applied to these models. As discussed in Section Methodology, the proposed PLoRA can be adapted to a broader range of neural networks as long as containing linear projectors.
Note that we do not want to limit the capability of our work to only sentiment analysis. Instead, our explorations could be extended to text-generative tasks and much broader applications as we introduced the text-generative paradigm of Flan-T5 to handle classification tasks. It sheds light on the promising future direction of PEFT and human-centered reasoning in large-scale language models.

\section{Conclusions} \label{Conclusions}
We proposed PLoRA, a human-centered PEFT approach that successfully demonstrated the effectiveness in enhancing transfer learning from pre-training to downstream tasks for PLMs. By adopting a PnP framework, PLoRA significantly improves its adaptative ability to the cold-start issues in real-world applications. The experiments conducted on diverse personalized sentiment analysis tasks validated the effectiveness and efficiency of our method. Moreover, this work not only contributes to the improvements of the performance of PLMs on text understanding tasks but also sheds light on future works, including the explorations of PLoRA on LayerNorm layer, CNN, and other hybrid components, and the service extension of applications.

\section*{Acknowledgments}
This work was supported by the National Natural Science Foundation of China (NSFC) under Grant Nos. 61966038 and 62266051,
 the Yunnan Ten Thousand Talents Program and Yunling Scholars Special Project under Grant YNWR-YLXZ-2018-022,
 the Ministry of Science and Technology (MOST), Taiwan, ROC, under Grant No. MOST110-2628-E-155-002,
 and the Yunnan Postdoctoral Science Foundation under Grant No. C615300504048.
 The authors would like to thank the anonymous reviewers for their constructive comments.

\bibliography{references}
\bibliographystyle{aaai24}

\appendix
\section{Appendices}
\subsection{Related Work} \label{RW}
This section briefly reviews the related works, including PEFT and personalized sentiment analysis.
\subsubsection{Parameter-Efficient Fine-Tuning}
PLMs, such as GPT-4 \cite{Peng2023} and BERT \citep{Devlin2019}, are developed in great part by the technique of transfer learning \citep{Pan2010}. One of the earliest explorations was using pre-trained word embeddings, such as word2vec \citep{Mikolov2013c} and GloVe \citep{Pennington2014}, to improve the performance of NLP models. More recently, with the emergence of large-scale language models, the scope of transfer learning has extended remarkably with a paradigm of pre-training and fine-tuning \citep{Liu2023}.
Nevertheless, fine-tuning an entire PLM has grown computationally expensive and frequently unfeasible as the parameter count of PLM approaches trillions \citep{Houlsby2019}. In response, the emphasis has changed to in-context learning \citep{radfordlanguage,Brown2020}, in which the model is provided with prompts for a given task and returns in-context updates. Its shortcomings, such as processing the prompt each time the model makes a forecast, as well as its occasionally poor performance, make it a less preferable option.

Considering only fine-tuning some of the model parameters \citep{Houlsby2019}, the technique of PEFT can save computational resources and time compared to training the entire model from scratch. It achieves this efficiency by freezing some of the layers of the pre-trained model and only fine-tuning the last few layers that are specific to the downstream task. The original adapter modules are a special kind of submodule that can be inserted into PLMs and modify the hidden representation during fine-tuning \citep{Pfeiffer2020}. The model only needs to update the parameters in the adapters during fine-tuning while leaving the rest of the model parameters fixed. LoRA considers the low-rank intrinsic dimensionality of matrices parameters for downstream tasks and performs low-rank update \citep{Hu2021}. Prefix-tuning improves a brief continuous task-specific vector termed the prefix while maintaining the PLM parameters at their initial values \citep{Li2021b}. In prefix-tuning, the prefix is a set of free parameters that are trained alongside the PLM. The goal is to identify a context that directs the large-scale language models to produce text that accomplishes a specific task. Similarly, Prompt tuning only appends these vectors into input embedding \citep{Lester2021}. Furthermore, P-tuning uses trainable continuous prompt embeddings to enhance PLMs performance \citep{Liu2021a}. The continuous prompts can be optimized using a downstream loss function and a prompt encoder, which helps solve discreteness and association challenges.

\begin{table*}[t]
\centering
\begin{tabular}{|c|c|c|c|c|c|c|}
\hline
\textbf{Datasets} & Classes & Train   & Dev    & Test   & Users & \#Reviews/user \\ \hline
IMDB-A            & 10      & 101,778 & 12,715 & 13,133 & 783   & 162.99         \\ \hline
IMDB-B            & 10      & 7,975   & 990    & 1,119  & 229   & 44.03          \\ \hline
YELP-A            & 5       & 298,713 & 37,328 & 38,988 & 3,247 & 115.50         \\ \hline
YELP-B            & 5       & 42,184  & 5,249  & 5,907  & 1,213 & 43.97          \\ \hline
GDRD-A            & 5       & 128,476 & 16,087 & 16,114 & 1,646 & 97.62          \\ \hline
GDRD-B            & 5       & 4,924   & 616    & 617    & 127   & 48.48          \\ \hline
PPR-A             & 5       & 40, 000 & 5,000  & 4,999  & 500   & 100.00         \\ \hline
PPR-B             & 5       & 7,600   & 950    & 950    & 190   & 50.00          \\ \hline
\end{tabular}
\caption{Datasets in statistics.} \label{tab:4}
\end{table*}

\subsubsection{Personalized Sentiment Analysis}
Sentiment analysis \citep{Liu2012a,yuan2023encoding} aims to automatically determine the attitudes of users regarding a specific target natural language content. Personalized sentiment analysis is an HCTU task and typically employs personalized background information, such as users and products, across lengthy document evaluations to pinpoint precise sentiments. The majority of background data is gathered from social networks like Amazon and IMDb. Deep learning techniques have been suggested to conduct tailored sentiment analysis, and they have produced excellent results.

\begin{table*}[t]
\centering
\begin{tabular}{|c|c|c|c|cccc|}
\hline
\textbf{Method}          & Checkpoint name                      & \# of params             & \textbf{Datasets}                  & \multicolumn{1}{c|}{IMDB} & \multicolumn{1}{c|}{YELP} & \multicolumn{1}{c|}{GDRD} & PPR \\ \hline
\multirow{7}{*}{-}       & \multirow{7}{*}{-}                   & \multirow{7}{*}{-}       & Optimizer                 & \multicolumn{4}{c|}{AdamW}                                                              \\ \cline{4-8} 
                         &                                      &                          & Batch size                & \multicolumn{4}{c|}{16}                                                                 \\ \cline{4-8} 
                         &                                      &                          & $r$                         & \multicolumn{4}{c|}{$r_q$=$r_v$=64}                                                           \\ \cline{4-8} 
                         &                                      &                          & Alpha of $r$                & \multicolumn{4}{c|}{128}                                                                \\ \cline{4-8} 
                         &                                      &                          & Learning rate (full-shot) & \multicolumn{4}{c|}{2E-5}                                                               \\ \cline{4-8} 
                         &                                      &                          & Learning rate (few-shot)  & \multicolumn{4}{c|}{1E-4}                                                               \\ \cline{4-8} 
                         &                                      &                          & MIM type                  & \multicolumn{4}{c|}{MSE}                                                                \\ \hline
\multirow{2}{*}{BERT}    & \multirow{2}{*}{\texttt{base-uncased}}        & \multirow{2}{*}{113.04M} & PDropout ratio            & \multicolumn{1}{c|}{0.2}  & \multicolumn{1}{c|}{0.4}  & \multicolumn{1}{c|}{0.1}  & 0.1 \\ \cline{4-8} 
                         &                                      &                          & MIM ratio                 & \multicolumn{1}{c|}{2.5}  & \multicolumn{1}{c|}{2.5}  & \multicolumn{1}{c|}{3}    & 2   \\ \hline
\multirow{2}{*}{RoBERTa} & \multirow{2}{*}{\texttt{roberta-base}}        & \multirow{2}{*}{128.80M} & PDropout ratio            & \multicolumn{1}{c|}{0.2}  & \multicolumn{1}{c|}{0.4}  & \multicolumn{1}{c|}{0.1}  & 0.1 \\ \cline{4-8} 
                         &                                      &                          & MIM ratio                 & \multicolumn{1}{c|}{2.5}  & \multicolumn{1}{c|}{2.5}  & \multicolumn{1}{c|}{2}    & 3   \\ \hline
\multirow{2}{*}{Flan-T5} & \multirow{2}{*}{\texttt{google/flan-t5-base}} & \multirow{2}{*}{257.42M} & PDropout ratio            & \multicolumn{1}{c|}{0.2}  & \multicolumn{1}{c|}{0.4}  & \multicolumn{1}{c|}{0.1}  & 0.1 \\ \cline{4-8} 
                         &                                      &                          & MIM ratio                 & \multicolumn{1}{c|}{2.5}  & \multicolumn{1}{c|}{2.5}  & \multicolumn{1}{c|}{3}    & 3   \\ \hline
\end{tabular}
\caption{Used configurations for our proposed methods.} \label{tab:5}
\end{table*}

\begin{table*}[!t]
\begin{tabular}{|m{7cm}|c|c|c|c|}
\hline
\multicolumn{1}{|c|}{\bf Text}  & $\cal B$-PLoRA* & $\cal B$-PLoRA (FS) & $\cal B$-PLoRA (ZS) & $\cal B$-PLoRA w/o ZS \\ \hline
Sample   1: great food and great service, what else is there to say? will definitely   come back, thanks! (VP) & VP (\checkmark)   & VP (\checkmark)       & VP (\checkmark)       & VP (\checkmark)         \\ \hline
Sample   2: i recently went to the gallery to see the andy warhol exhibit. there was a   locals discount which i enjoyed --rrb-- also if you go on a wednesday its   only 8 for locals so if you want to see it i would suggest that day because   that is a great deal! i really enjoyed this exhibit. i think andy warhol had   some great paintings and it was fun to learn about them. when you go into the   exhibit you are given a device that has recordings for each art piece so you   are able to learn about each piece as you go through the exhibit. it was kind   of small but it was the perfect size for me. since i'm not an art fanatic and   just wanted to see the main pieces it was great. for anyone who loves art i   would recommend this exhibit. it was nice to be able to see something   different on the vegas strip. this exhibit is here until october so you have   plenty of time to check it out if you wish to do so. (P) & VP ($\times$)   & P (\checkmark)        & VP ($\times$)       & VP ($\times$)         \\ \hline
Sample   3: i ate here often when i lived in the area. cheap, good pizza. has anyone   else noticed that they are not using a brick oven to cook their pizzas? (P)   & P (\checkmark)    & P (\checkmark)        & P (\checkmark)        & VP ($\times$)         \\ \hline
Sample 4: my boyfriend and i ate here because he had a   groupon. normally i don't dig italian but i tried to stay open... the menu is   great and all but, man, salty food! my minestrone soup was good, but both my   salad lrb who salts a salad??!!! –rrb-- and my boyfriend's chicken parmigiana   dish were so heavily salted you could barely taste any distinct flavors. the   service was good, and all else but man, i couldn't drink enough fluid the   rest of the night. (Ne) & Ne (\checkmark)   & Ne (\checkmark)       & VN ($\times$)       & VN ($\times$)         \\ \hline
Sample 5: great   soccer fields, a small swing area for kids and lot of shaded area. only   negative is bathrooms are dirty. (Ne) & VP ($\times$)   & VP ($\times$)       & VP ($\times$)       & P ($\times$)          \\ \hline
\end{tabular}
\caption{Example reviews from dev sets of YELP-B and the corresponding predictions of various methods. Very Negative (VN), Negative (N), Neutral (Ne), Positive (P), Very Positive (VP) for ratings ranging from 1 to 5, respectively.} \label{tab:6}
\end{table*}

To learn representations, the earlier methods pioneer traditional neural networks models, including CNN \citep{Kim2014}, LSTM \citep{Wang2018}, and GRU \citep{Zulqarnain2020}. To integrate user and product information into textual representations, recent works typically consider them as external knowledge features and designed knowledge injection methods for incorporation. Initially, \citet{Tang2015a} introduced a matrix-based method to project word embeddings into personalized ones, followed by a representation concatenation method for classification. Based on the hierarchical attentive network (HAN), \citet{Chen2016} and \citet{Wu2018} proposed attention-based injections that inject personalized information into attention maps. Recently, \citet{Amplayo2019} improved matrix-based methods and proposed the chunk-wise importance matrix (CHIM) method to effectively and efficiently incorporate user and product information into various sections of sentiment models.

Recent researches aim to inject personalized knowledge into PLMs for better performance with the paradigm of pre-training and fine-tuning. For example, MA-BERT introduced six personalized-transformer layers stacking on pre-trained BERT and achieved effective improvements \citep{Devlin2019}. UserAdatper used a prompt tuning method to incorporate user information and proposed a few-shot learning strategy for newly registered users with a few data samples \citep{Zhong2021}. To make current personalized models capable of providing versatile services for unseen users, \citet{Zhang2023} propose a switch knowledge distillation method to narrow the distances between personalized and generalized representations of reviews.

Our method of PLoRA, discussed in detail in Section Methodology, offers a spectrum of advantages over related PEFT and PKI works. In comparison of existing personalized models with sophisticated structure, PLoRA is more efficient as PLoRA simply yet effectively combines LoRA and PKI. Similar to LoRA, PLoRA does not increase sequence lengths when it handles text input and does not have additional inference latency in comparison of other PEFT methods. Furthermore, we introduce a PnP framework for PLoRA where PLoRA is easy-to-deploy and capable of adapting to complex cold-start scenarios in practice.

\subsection{Dataset Details}
{\bf IMDB} is a wide collection of movie reviews from IMDb website where recommendation ratings for each review ranges from 1-10 stars. The current version is introduced in \citep{Zhong2021} and contains two main parts (denoted A and B) for full-shot and few/zero-shot learning scenarios.

\noindent{\bf YELP} is introduced in \citep{Zhong2021} and a collection of restaurant reviews where review ratings ranges from 1-5 stars. It also contains A and B parts for full-shot and few/zero-shot learning scenarios.

\noindent{\bf GDRD} \footnote{https://mengtingwan.github.io/data/goodreads} contains book reviews collected from Goodreads websites, corresponding with 1,773 users \citep{Wan2018}. The sentiment rating for each review ranges from 1-5 stars.

\noindent{\bf PPR} \footnote{https://www.kaggle.com/shuyangli94/food-com-recipes-and-user-interactions} is a personalized review dataset that contains food recipe reviews collected from GeniusKitchen websites and gathered from 690 users \citep{Majumder2019}. The sentiment rating for each review ranges from 1-5 stars.

Similar to IMDB and YELP, our introduced GDRD and PPR personalized datasets are also divided into A and B parts. More statistics of these datasets are reported in Table 4.

\subsection{Hyperparameter Used in Experiments}
For optimization setup, we train all models using AdamW optimizer with early stopping strategy of 5 epoch patience. Test models are selected on dev datasets where the $Acc$ score is considered as a monitor metric following the previous work \citep{Tang2015a,Zhang2021,Wu2018}. We sweep learning rate for full/few-shot learning scenarios, batch sizes, and configurations for PLoRA. We report average experimental results over 5 runs with different seeds. Notably, the computational costs exponentially increase with input length \citep{Beltagy2020} and most of review lengths in IMDB and YELP statistically exceed 512 tokens that is the maximum input length for BERT, RoBERTa and T5. Therefore, we use a truncate strategy to concatenate the first 256 and the last 256 tokens when it meets reviews over 512 tokens. See the hyperparameters used in our models in Table 4.

\subsubsection*{BERT \& RoBERTa}
Both BERT and RoBERTa are Transformer encoder-based PLMs and are pre-trained with mask language model tasks. To generate the sentiment score of a sentence, we extract hidden states of start-specific tokens ({\texttt{[CLS]}} for BERT and \texttt{<s>} for RoBERTa) as the text representations that are then fed into the classifier for prediction.

\subsubsection{Flan-T5}
The backbone of Flan-T5 is T5, which is structured in an encoder-decoder Transformer and pre-trained with text-to-text tasks. With the diagram shift transferring classification to text generation tasks, we introduce prefix textual instruction “Review: $\bf x$” for text inputs x and corresponding output format of “The sentiment score is $y$” for a sample $({\bf x}, y)$.

\subsection{Optimization in PnP}
\subsubsection{Initialization in PLoRA}
We rethink the Eq. (3) as 
\begin{equation}
\begin{aligned}
  h' &= g(h,p;W,W_{{\rm{task}}}^{{\rm{in}}},W_{{\rm{task}}}^{{\rm{out}}},W_{{\rm{person}}}^{{\rm{in}}},W_{{\rm{person}}}^{{\rm{out}}}) \cr 
     &= {g_1}(h;W)  \cr
     &\circ {g_2}(h;W_{{\rm{task}}}^{{\rm{in}}},W_{{\rm{task}}}^{{\rm{out}}})  \cr
     &\circ {g_3}(p;W_{{\rm{person}}}^{{\rm{in}}},W_{{\rm{person}}}^{{\rm{out}}}) \cr
\end{aligned}\,,
\end{equation}
where $W$ represent the matrix weight originating from PLMs; $h$, $p$ and $h'$ denote the text input, the personal knowledge and the output of PLoRA, respectively; $W_{{\rm{task}}}^{{\rm{in}}}$ and $W_{{\rm{task}}}^{{\rm{out}}}$ are trainable parameters in the task-specific part; and $W_{{\rm{person}}}^{{\rm{in}}}$ and $W_{{\rm{person}}}^{{\rm{out}}}$ mean the human-centered part. ${g_1}( \cdot )$, ${g_2}( \cdot )$, and ${g_3}( \cdot )$ are functions of PLM, LoRA, and PKI forward propagation and $ \circ $ is corresponding composition operator.

Because both $p$ and $W_{{\rm{task}}}^{{\rm{out}}}$ are initialized as zero, in the first step for either downstream task or new-participated user, we have:
\begin{equation} \label{eq:7}
\begin{cases}
\begin{aligned}
g(h) &= {g_1}(h) & {g_2}(h) = 0;{g_3}(p) = 0 \cr
g(h) &= {g_1}(h) \circ {g_2}(h) & {g_3}(p) = 0 \cr
\end{aligned}
\end{cases} \,.
\end{equation}
This indicates that, in the first training step, all inputs and outputs of both PLoRA are consistent with they would be in the original PLM comments. In this way, the capability, functionality, and result quality of adopted blocks is impeccably preserved and further optimization will become as fast as fine-tuning.

\subsubsection{Optimization for PLoRA}
A brief gradient calculation of PLoRA is deduced as,
\begin{equation}
\begin{aligned}
  {{\partial g(h,p)} \over {\partial W_{{\rm{task}}}^{{\rm{in}}}}} &= hW_{{\rm{task}}}^{{\rm{out}}} \cr 
  {{\partial g(h,p)} \over {\partial W_{{\rm{person}}}^{{\rm{in}}}}} &= pW_{{\rm{task}}}^{{\rm{out}}} \cr 
  {{\partial g(h,p)} \over {\partial p}} &= W_{{\rm{person}}}^{{\rm{in}}}W_{{\rm{task}}}^{{\rm{out}}} \cr 
  {{\partial g(h,p)} \over {\partial W_{{\rm{person}}}^{{\rm{out}}}}} &= {{\partial g(h,p)} \over {\partial W_{{\rm{task}}}^{{\rm{out}}}}} = hW_{{\rm{task}}}^{{\rm{in}}} + pW_{{\rm{person}}}^{{\rm{in}}} \cr
\end{aligned} \,.
\end{equation}
We can see that, although $p$ and $W_{{\rm{task}}}^{{\rm{out}}}$ are zeros at the first training step, corresponding gradients are non-zero and make zero-initialized parameters optimized in the future steps. Moreover, the gradients to $W_{{\rm{person}}}^{{\rm{out}}}$ is calculated with both task-specific and personalization-specific forward propagation. At the same time, $W_{{\rm{person}}}^{{\rm{out}}}$ also helps calculations of the gradients to $W_{{\rm{task}}}^{{\rm{in}}}$ and $W_{{\rm{person}}}^{{\rm{in}}}$, which facilitates the fusion of task-specific and personalization-specific optimization for PLoRA.

\subsection{Additional Experiments}
Some cases sampled from the dev dataset in YELP-B and corresponding predictions from variants of $\cal B$-PLoRA were shown in Table 6.

Sample 1. This is a straightforward and simple restaurant review for Very Positive reviews. Thus, all models make correctly predictions.

Sample 2. This review conveys Positive sentiment for the user (also the writer). $\cal B$-PLoRA* fails to predict the correct rating due to the immature semantics of the user learned from a handful of data for the user. By contrast, $\cal B$-PLoRA (FS) correctly predict the golden sentiment due to its robust few-shot learning performance. Obviously, without the consideration of the semantics of the user, $\cal B$-PLoRA w/ or w/o ZS fails to accurately predict the review sentiment.

Sample 3. This review is similar to the 2nd sample that conveys Positive sentiment. Only $\cal B$-PLoRA w/o ZS fails to locate the correct sentiment score. This sample show the importance of zero-shot learning performance in our proposed methods.

Sample 4: This review in fact expresses a Neutral sentiment where models with PKI successfully predict the correct sentiment while those without PKI fails to make accurate predictions. This review shows the importance of personalized knowledge for sentiment analysis.

Sample 5: All models, regarded as whether few/zero-shot learning strategies they use, predict (Very) Positive sentiment while the ground-truth sentiment is Neutral. This is a difficult example where the overall sentiment is subtly expressed.

\end{document}